\documentclass[letterpaper, 10 pt, conference]{ieeeconf}  
\usepackage{cite} 
\usepackage[table]{xcolor}
\definecolor{myorange}{HTML}{FF9D00}
\definecolor{mypink}{HTML}{FF00B2}
\definecolor{mypurple}{HTML}{00B227}

\definecolor{myred}{HTML}{00F6FF}
\definecolor{mygreen}{HTML}{8800FF}
   
\usepackage{siunitx}
\usepackage{balance} 
\usepackage{bbold}
\usepackage{ulem}
\usepackage{soul}
\usepackage{float}
\usepackage{hyperref}
\usepackage{multirow}
\usepackage{xcolor,colortbl}
\definecolor{ashgrey}{rgb}{0.7, 0.75, 0.71}
\definecolor{gainsboro}{rgb}{0.86, 0.86, 0.86}
\usepackage{graphicx}
\usepackage{subcaption}
\usepackage{comment}
\usepackage{dsfont}
\usepackage{mathtools}
\usepackage{array}
\usepackage{amsmath}
\usepackage{amssymb}
\usepackage{booktabs}
\usepackage{soul}
\usepackage{multirow}
\usepackage{adjustbox}
\usepackage{hyperref}
\usepackage{hyperref}
\usepackage{graphicx}
\usepackage{amsmath}
\usepackage{amssymb}
\usepackage{booktabs}
\usepackage{comment}
\usepackage{soul}

\usepackage{tikz}
\usetikzlibrary{arrows.meta,positioning,fit,calc}

\IEEEoverridecommandlockouts                              
           
\hypersetup{
    colorlinks=true,
    urlcolor=green,
    linkcolor=green,
    citecolor=green
}                                               

\title{\LARGE \bf
Towards Socially Compliant Navigation in Deep Reinforcement Learning via Proxemics-Based Reward Modeling
}

\author{%
Takieddine Soualhi$^{1,\ast}$,
Jacques Saraydaryan$^{2}$,
Laetitia Matignon$^{3}$%
\thanks{$^{\ast}$ Corresponding author. }%
\thanks{$^{1}$ Inria, CITI Lab, INRIA--INSA Chroma Team, Villeurbanne, France.
{\tt\small takieddine.soualhi@inria.fr}}%
\thanks{$^{2}$ CPE Lyon, CITI Lab, INRIA--INSA Chroma Team, Villeurbanne, France.
{\tt\small jacques.saraydaryan@cpe.fr}}%
\thanks{$^{3}$ Université Lyon 1, INSA Lyon, CNRS, LIRIS, UMR 5205, Villeurbanne, France.
{\tt\small laetitia.matignon@univ-lyon1.fr}}%
\thanks{This work was funded by the French National Research Agency through the SOLARNav project (ANR-23-DMRO-0018).}
}
\begin{document}

\maketitle
\thispagestyle{empty}
\pagestyle{empty}

\begin{abstract}
Developing effective robot navigation methods in crowded environments is essential for real-world applications. Although recent deep reinforcement learning (DRL) methods have improved navigation performance in crowded environments, they often focus primarily on task-centric objectives and underrepresent social compliance objectives. In this paper, we introduce a novel proxemics-based reward formulation for DRL social navigation that provides a dense, interpretable social learning signal while maintaining navigation efficiency. Our approach models each human’s personal space as a radial Gaussian-mixture field derived from Hall’s proxemics theory and computes a robot-centric local cost over the robot’s field of view. We integrate the proposed reward into established DRL navigation methods and evaluate it in simulation across multiple crowd scenarios, reward baselines, and crowd densities using both navigation metrics and social metrics. Results show that the proposed reward consistently improves social metrics in simulation while maintaining competitive navigation performance relative to the compared reward models. \\
Project page: \ \url{https://drl-proxemics.github.io/}
\end{abstract}


\section{Introduction}

Robots are increasingly expected to operate in crowded environments such as malls, hospitals and airports. In these settings, navigation is not merely a geometric or dynamic problem of reaching a goal without collisions: robots must also move in ways that humans perceive as safe, predictable, smooth and minimally intrusive \cite{mavrogiannis_core_2023}. This requirement is commonly framed within the problem of social navigation, where success depends jointly on navigation task efficiency and social compliance. A widely adopted conceptual framework for social navigation is the set of principles articulated by Francis et al.\cite{francis2025principles}, which emphasize safety and comfort for surrounding agents, adherence to social norms, and legible motion through smooth and intentional trajectories. These principles make explicit that collision avoidance is necessary but insufficient, a robot can remain collision-free while still producing behaviors that humans find socially inappropriate.

Existing approaches to social navigation are often grouped into (i) optimization and model-based methods and (ii) learning-based methods \cite{mavrogiannis_core_2023}. Optimization-based approaches typically compute trajectories by relying on hand-crafted mathematical models subject to feasibility constraints. In addition, grounded in Hall’s proxemics theory which studies how humans perceive and regulate interpersonal space during social interaction \cite{hall1966hidden,truong2017approach}, these methods can encode comfort-related criteria such as interpersonal distance zones directly into local navigation costmaps, offering interpretability and, in some cases, strong theoretical guarantees. However, their performance can degrade when assumptions about human motion, scene structure, or interaction patterns are violated, and they may require substantial manual tuning to remain robust across diverse crowd configurations \cite{le2024social}.

Learning-based methods, and in particular deep reinforcement learning (DRL), reduce reliance on explicit modeling by training control policies in simulation. This enables policies to exploit rich observations and to learn interaction patterns that can be difficult to handcraft \cite{Chen2017CADRL}. However, the central difficulty is that DRL policies are strongly shaped by the reward function and the evaluation protocol. If comfort and social acceptability are not explicitly represented, they are typically optimized only indirectly, to the extent that the reward design or training data capture them. While, prior work has sought to address this through reward modelling, including proxemic considerations \cite{Flogel2024SociallyIntegrated, maddumage2025relative}. The existing reward formulations are often sparse, rely on global state information rather than local perception, or lack principled grounding in Hall's proxemic theory. Consequently, DRL-based navigation can achieve high navigation success rates while still producing behaviors that violate personal space, exhibit high movement jerk, or appear abrupt. 
\begin{figure}[t]
    \centering

\includegraphics[width=0.91\columnwidth]{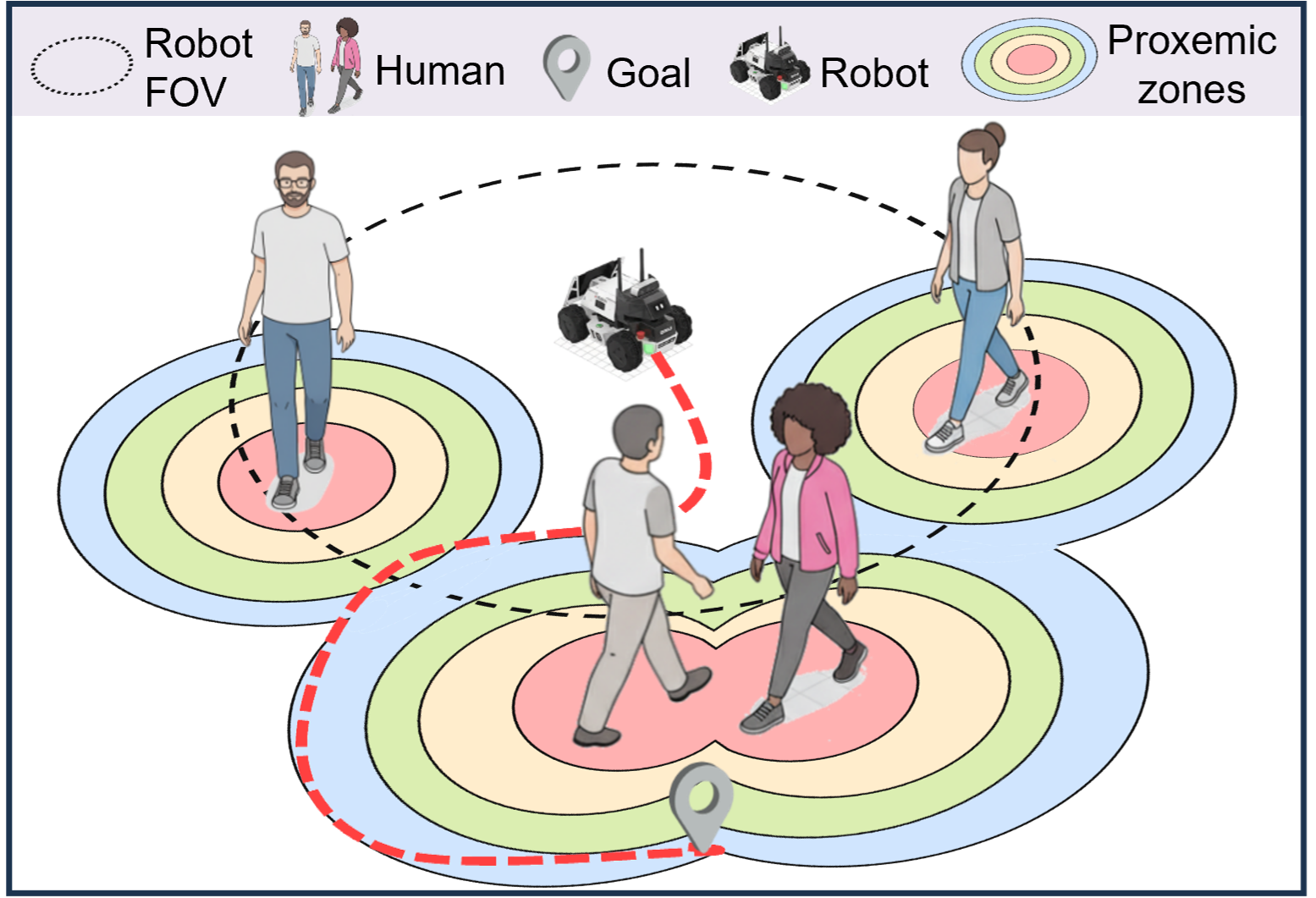}

\caption{Illustration of the addressed DRL social navigation problem. The robot must reach a goal while navigating among humans and respecting proxemic zones. The dashed red trajectory illustrates a socially compliant path.}
    \label{fig:overview}
\end{figure}

This paper addresses the gap between collision-free navigation and human-aware navigation in DRL. Rather than proposing a new policy architecture, we focus on reward modeling and ask whether a proxemics-grounded signal can consistently improve human-aware behavior across different DRL navigation backbones (Figure~\ref{fig:overview}). To this end, we introduce a proxemics-based reward formulation that represents each pedestrian’s personal space as a radial Gaussian-mixture field, and computes a robot-centric local proxemic cost within the field of view (FOV) of the robot. The proposed reward model encourages the robot to minimize proxemic intrusion while maintaining navigation efficiency, thereby promoting smoother, more legible interactions in dense crowds. Beyond reward modeling, we emphasize that comfort-aware navigation must be assessed explicitly rather than relying on navigation metrics alone as proxies. We evaluate the proposed formulation across established DRL navigation methods and reward baselines using a set of task and comfort-oriented metrics. In a nutshell, the main contributions of this paper are:
\begin{itemize}
 \item A novel robot-centric, proxemics-based reward model grounded in Hall’s theory of interpersonal distance.
\item Validation across two DRL navigation methods, showing consistent social metrics gains with competitive performance relative to the compared reward models.
\item A systematic analysis of key reward design factors providing insights for reward modeling in social navigation.
\end{itemize}

The remainder of this paper is organized as follows. Section~\ref{sec:related_work} reviews related work. Section~\ref{sec:method} formalizes the social navigation task and presents the reward model within a DRL framework. Section~\ref{sec:experimental_setup} describes the experimental setup and evaluation protocol. Section~\ref{sec:experimental_results} reports results, comparisons to existing reward formulations, and ablation analyses. Section~\ref{sec:conclusion} concludes and outlines future directions.

\section{Related Work}
\label{sec:related_work}
\subsection{DRL-based social navigation}

Early DRL-based methods focused on single robot navigation in open environments without static obstacles, showing that dynamic obstacle avoidance can be achieved by training models in simulation \cite{GA3C-CADRL}. These approaches typically encode interactions in the reward function, but their performance often degrades in dense crowds. To address this, later work explicitly modeled robot-crowd interactions using interaction graphs that capture human-human (HH) and human-robot (HR) relationships, where nodes represent humans or robots and edges encode interactions. Recent methods such as ST2 \cite{yang2023st}, AttnGraph \cite{liu2023attngraph}, and NaviSTAR \cite{wang2023navistar} leverage graph-based architectures such as graph convolutional networks and graph attention networks to learn richer representations of these interactions. Building on single robot social navigation, another line of research extends DRL to multi-robot settings, using graph-based models to capture robot–robot (RR) interactions and enable implicit coordination in crowds. Zhou et al. \cite{zhou2025her} introduced HeR-DRL, which uses a heterogeneous relation graph to model HH, HR, and RR interactions. Escudie et al. \cite{multisoc} proposed MultiSoc, a graph-attention and multi-agent DRL approach that generalizes to both single and multi-robot scenarios.

\subsection{Reward modeling in DRL-based social navigation}
Social compliance in DRL-based social navigation is typically enforced through a combination of explicit constraints and reward modelling. To address the safety concerns in DRL-based social navigation, a common approach is to augment policy learning with geometric/kinematic safety priors. For instance, DRL-VO integrates a velocity-obstacle (VO) based term into the reward to bias the learned policy away from imminent collisions while preserving throughput in dense crowds \cite{xie2023drl}. Zhu et al. \cite{Zhu2025ConfidenceAware} introduced a constrained DRL method that incorporates dynamic distance constraints into the training objective aiming to avoid both unsafe behavior and overly conservative policies. On the other hand, several works focused on going beyond collision avoidance, by designing  reward models that constrain the robot to specific social behaviours \cite{Chen2017SACADRL}. Qiu et al. \cite{Qiu2022InteractionCapacity} proposed a reward model that enables interactive robot approaches, where robots learn to negotiate passage (e.g., by beeping) to prevent the “freezing” problem in crowds. Nishimura et al. \cite{Nishimura2020L2B} study the safety/efficiency trade-off and propose a reward function that promotes a better balance by penalizing both frequent active path clearing and passive collision avoidance.
Flögel et al. \cite{Flogel2024SociallyIntegrated} extend the formulation of Nishimura et al. \cite{Nishimura2020L2B} into a broader socially integrated DRL navigation framework by incorporating person-specific threshold radii within a social reward formulation. Maddumage et al. \cite{maddumage2025relative} proposed a relative-velocity reward model that penalizes high closing relative velocities between the robot and nearby humans, encouraging safer social navigation.

\subsection{Discussion}
The aforementioned literature shows substantial progress in DRL-based social navigation, but also a key gap. Much prior work focuses on DRL architectures and training formulations, typically optimizing navigation-centric objectives. As a result, learned policies may still produce abrupt speed changes or intrude on personal space. Existing reward modeling approaches for comfort also have limitations: proxemic comfort is often represented via sparse, threshold-based signals that are hard to learn through DRL \cite{Nishimura2020L2B,Flogel2024SociallyIntegrated}, while kinematic surrogates such as relative velocity provide only indirect supervision \cite{maddumage2025relative}. Compared to socially integrated frameworks like Flögel et al. \cite{Flogel2024SociallyIntegrated}, our focus is narrower and complementary: we isolate proxemics-based reward modeling as a modular component to study the impact of reward design itself.
Evaluation protocols further tend to emphasize task success and efficiency, with limited focus on human comfort. To address this gap, we propose a proxemics-based reward that models personal space via a radial Gaussian mixture field and computes a local cost over the robot’s FOV. We also evaluate multiple established DRL methods and recent reward models \cite{maddumage2025relative,Nishimura2020L2B} using both task and comfort metrics, aligning with recent calls to move beyond purely task-centric evaluation \cite{francis2025principles}.

 \begin{figure*}[!ht]
     \centering
       \includegraphics[width=1.0\linewidth]{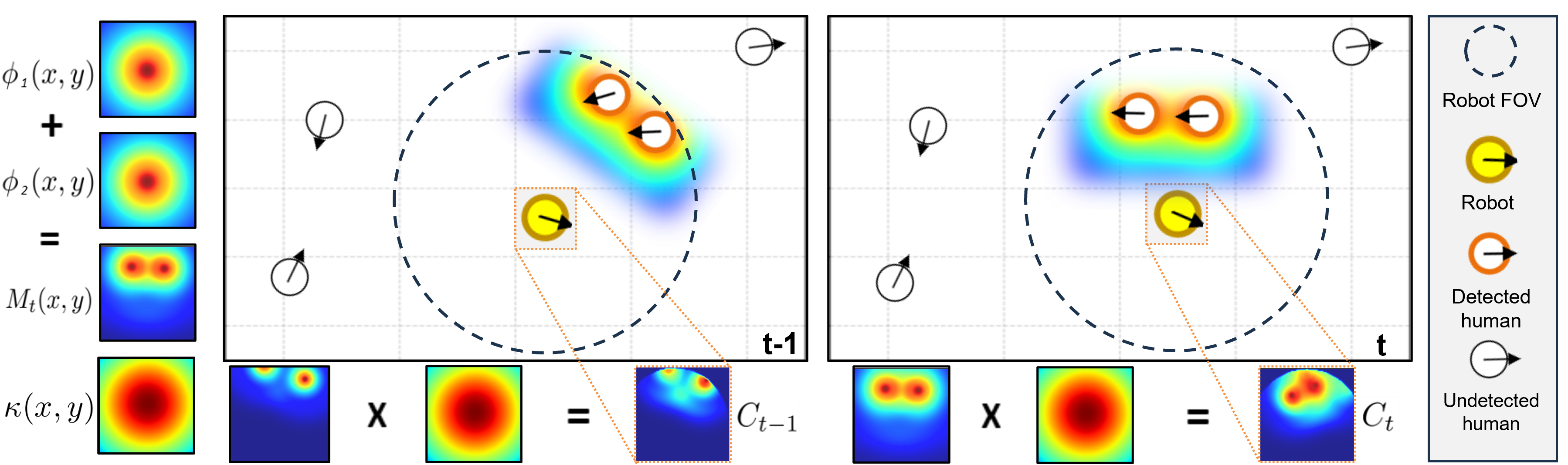}
     \caption{Illustration of the proposed reward model. Detected humans generate individual proxemic fields $\phi_j(x,y)$, which are aggregated into a social cost map and locally weighted by a robot-centered kernel $\kappa(x,y)$ to compute $C_{t-1}$ and $C_t$. The proxemic reward $r_{\mathrm{prox},t}$ is then computed from their temporal difference.}
     \label{fig:reward}
 \end{figure*}
\section{Proxemics for DRL-based Social Navigation}
\label{sec:method}
In this section, we formulate the socially compliant robot navigation problem and introduce our proposed reward model.

\subsection{Problem statement and formulation}

Consider a mobile robot operating in a scene populated by a human crowd $N_h$. Starting from an initial state, the robot must reach a goal position $g$ using only its perception system and no map is available. Our primary objective is safe and comfortable navigation. The robot has to reach $g$ while avoiding collisions and maintaining a comfortable separation from nearby humans. Within RL literature, this problem can be formulated as a POMDP \cite{kaelbling1998planning} defined by $\langle \mathcal{S}, \mathcal{A}, \mathcal{T}, \mathcal{O}, \mathcal{Z}, \mathcal{R} \rangle$. Here,  $\mathcal{S}$ denotes the state space which includes robot and humans positions and velocities. At each timestep $t$, the robot receives a local observation $o_t \in \mathcal{O}$, executes an action $a_t \in \mathcal{A}$, and receives a  scalar reward $r_t$, while $\mathcal{T}$ (the state transition model) and $\mathcal{Z}$  (the observation model) are assumed to be unknown. The objective is to learn a policy $\pi(a_t \mid o_t)$ that maximizes the expected discounted return.

For a social navigation problem, the robot receives an observation $o_t = \{w_t, H_t\}$ consisting of two components: (i) an intrinsic information vector $w_t$, which encodes the robot's internal state and goal, and (ii) a set of detected humans $H_t$ within its perception FOV. The intrinsic information is expressed as ${w_t = [p_t, v_t, g, \theta_t, \rho_t]}$, where $p_t$ denotes the robot’s current position, $v_t$ its velocity, $g$ its goal coordinates, $\theta_t$ its yaw angle, and $\rho_t$ its radius. The detected human set is given by $H_t = \{p_t^j\}_{j \in \mathcal{J}_t}$, 
with $\mathcal{J}_t \subseteq \{1, \dots, N_h\}$, where $p_t^j=(x_t^j,y_t^j)$ is the 
position of human $j$ expressed in the robot’s local frame. Humans are detected only within the robot’s perception FOV, defined by its sensing range and angular aperture. In this formulation,  we assume access to only detected positions, rather than full state estimates, to better match realistic deployments where off-the-shelf detectors can reliably provide relative human positions, while velocities and long-horizon predictions may be more complex to obtain.

As for kinematics, we adopt a holonomic kinematic model in which, at each instant $t$, the robot executes a continuous action $a_t = (v_{x,t}, v_{y,t})$
where $v_{x,t}$ and $v_{y,t}$ denote the planar velocity components along the robot’s local $x$ and $y$ axes, respectively. This formulation allows the robot to move in any direction on the plane and is consistent with common assumptions in prior work \cite{liu2023attngraph,yang2023st}.

\subsection{Methodology}

Our proposed reward model is designed to promote socially compliant navigation by integrating proxemics. This reward is computed from locally perceived human positions within the robot’s FOV, making it compatible with partial observability and realistic perception constraints. First, for each human $j \in \{1, \cdots, N_h\} $ with position $p_t^j$, we define a radially symmetric proxemic field $\phi_j(x,y)$ that assigns higher cost to regions closer to the human and smoothly decays with distance. The field is implemented as an isotropic Gaussian mixture over the radial distance:
\begin{equation}
\phi_j(x,y)
=
\sum_{k\in\{\text{int},\text{per},\text{soc},\text{pub}\}}
A_k\,
\exp\!\left(
-\frac{\big(d_j(x,y)-\mu_k\big)^2}{2\sigma_k^2}
\right).
\label{eq:1}
\end{equation}

where $d_j(x,y)=\|(x,y)-p_t^j\|_2$ is the Euclidean distance from location $(x,y)$ in the map to human $j$, centered at the position $p_t^j$. Following Hall’s proxemic theory and prior work on model-based social navigation \cite{truong2017approach}, we associate the four interpersonal zones (intimate, personal, social, public) with mixture components $k\in\{\text{int},\text{per},\text{soc},\text{pub}\}$: $\mu_k$ specifies the characteristic radius of the zone, $\sigma_k$ controls its spatial spread, and $A_k$ is a scalar that weights its relative importance. It is important to note that we adopt this isotropic form based on the finding that it performs better for our DRL-based navigation setting, unlike the anisotropic fields commonly used in model-based social navigation. 

Given the per-human proxemic fields $\{\phi_j(x,y)\}_{j\in \mathcal{J}_t}$, we define the social cost map:
\begin{equation}
M_t(x,y) = \mathrm{Norm}(\sum_{j \in \mathcal{J}_t} \phi_j(x,y)).
\end{equation}
Here, $\mathrm{Norm}(C) = 1 - e^{-C}$ maps $[0, +\infty)$ to $[0, 1)$, 
assigning higher cost to proxemic intrusions and increasing naturally where personal spaces overlap, ensuring that the aggregated cost remains 
bounded and interpretable regardless of crowd density. Although we do not explicitly model groups, this 
superposition implicitly captures group-like structure: overlapping fields 
raise the cost between nearby humans, discouraging the robot from cutting 
through congested areas. In practice, $M_t$ is approximated by sampling it on an $N_x \times N_y$ grid, yielding the cost map matrix $\mathbf{M}_t \in \mathbb{R}^{N_x \times N_y}$.

To capture the spatial footprint of the robot’s influence around its body and immediate vicinity, we define a robot-centered kernel $\kappa(x,y)$, chosen here as a radial Gaussian:
\begin{equation}
\kappa(x,y)
=\exp\!\left(
-\frac{x^2+y^2}{2\sigma_r^2}
\right).
\end{equation}
Here, $\sigma_r$ controls the spatial extent over which the robot contributes to proxemic intrusion. The kernel is evaluated over the robot’s instantaneous FOV $\Omega_t$, defined by sensing range and angular aperture. In practice, we evaluate $\kappa$ on a grid over $\Omega_t$ to form the kernel matrix $\mathbf{K}_t \in \mathbb{R}^{N_x \times N_y}$.


At each timestep, we compute a scalar local proxemic intrusion cost by integrating the product of the robot kernel and the aggregated human proxemics:
\begin{equation}
\begin{aligned}
C_t
&=
\iint_{\Omega_t}
\kappa(x,y) M_t(x,y)\, dx \, dy,\\[0.5em]
&\approx
\langle \mathbf{K}_t,\mathbf{M}_t\rangle_F, \\[0.5em]
&=
\sum_{m=1}^{N_x}\sum_{n=1}^{N_y}
(\mathbf{K}_t)_{m,n}\,(\mathbf{M}_t)_{m,n}.
\end{aligned}
\end{equation}
\noindent where $\langle\cdot,\cdot\rangle_F$ denotes the Frobenius inner product, and the discrete kernel is normalized such that $\sum_{m,n}(\mathbf{K}_t)_{m,n}=1$, making $\langle \mathbf{K}_t,\mathbf{M}_t\rangle_F$ a kernel-weighted average of proxemic cost over $\Omega_t$. The idea behind this formulation is to quantify the spatial overlap between the robot’s local neighborhood and human-induced proxemic cost by computing a weighted cost over the instantaneous FOV. Intuitively, $C_t$ increases when the robot’s footprint overlaps high-cost regions near humans and decreases as the robot moves toward lower-cost areas. It is important to note here, because the sum is restricted to $\Omega_t$, humans outside the FOV do not contribute to $C_t$. 
Finally, we define our proposed proxemics-based reward term as the temporal change in the proxemic cost:
\begin{equation}
r_{\text{prox},t} = -\lambda \cdot \Delta C_t.
\label{eq:5}
\end{equation}
where $\Delta C_t = C_t - C_{t-1}$ and $\lambda$ is a weighting coefficient. Actions that increase proxemic intrusion are penalized, while actions that reduce intrusion are rewarded. We use $\Delta C_t$ (rather than $C_t$) following \cite{ibrahim2024comprehensive}, which notes that temporal-difference shaping can yield lower-variance learning signals and improve advantage estimation. Because this term is used only for shaping and is to be combined with the navigation reward, it does not encourage stalling at high proxemic cost.

We use a social navigation reward structure \cite{liu2023attngraph} in which the terminal rewards are fixed. Only the goal progress and social interaction reward terms vary with time. The reward at time step $t$ is defined as
\begin{equation}
r_t =
\begin{cases}
r_c & \text{if collision}, \\
r_s & \text{if goal reached}, \\
r_{\text{pot},t} + r_{\text{soc},t} & \text{otherwise}.
\end{cases}
\label{eq:6}
\end{equation}
where $r_c$ is a negative terminal penalty applied upon collision and $r_s$ is a positive terminal reward when the robot reaches the goal. The potential based shaping term encourages progress toward the goal and is given by ${r_{\mathrm{pot},t} = d_{g,t-1} - d_{g,t}}$, where $d_{g,t}$ denotes the Euclidean distance between the robot and its goal at timestep $t$. 
The term $r_{\text{soc},t}$ denotes the social interaction reward and is the only component changed across reward variants (Sec~\ref{sec:rewards}). 
\section{Experimental Setup}
\label{sec:experimental_setup}

\subsection{Implementation Details}

\paragraph{Simulation}
We use the CrowdNav simulator \cite{liu2023attngraph}. Following prior work~\cite{liu2023attngraph}, we assume a $360^\circ$ perception FOV with a $5\mathrm{m}$ range, so the kernel domain $\Omega_t$ is a disk of radius $R_{\mathrm{FOV}}=5\mathrm{m}$ centered at the robot in its local frame (see Section~\ref{sec:fov} for FOV ablations). We consider scenarios with $N_h = 15$ humans, simulated using ORCA \cite{vandenberg2011reciprocal}. Prior work suggests DRL policies are reasonably robust to the simulator choice (e.g. social forces), so the main trends should not depend strongly on it (see Section~\ref{sec:human} for ablations wrt $N_h$). For reward computation, $\Omega_t$ is discretized at a resolution of 
$0.1\,\mathrm{m}$ per cell, yielding $N_x = N_y = 101$.
\paragraph{Scenario} 
We consider two evaluation scenarios, Circle Crossing and Corridor. In Circle Crossing, a set of humans is initialized approximately on a circle, and each human is assigned a goal on the opposite side, inducing crossing trajectories through the center region. In Corridor, humans and the robot navigate in a corridor like formation toward assigned goals. In both scenarios, whenever a human reaches its goal, a new goal is immediately resampled at random. The robot starts from a random position with an independently sampled random goal. The episode ends when it either reaches the goal or collides.
\begin{table}[!h]
\centering
\scriptsize
\begin{tabular}{lccc}
\toprule
Zone & $\mu_k$ (m) & $\sigma_k$ (m) & $A_k$ \\
\midrule
Intimate (int) & 0.00 & 0.20 & 0.50 \\
Personal (per) & 0.45 & 0.30 & 0.30 \\
Social (soc) & 1.00 & 0.50 & 0.15 \\
Public (pub) & 2.60 & 0.90 & 0.05 \\
\bottomrule
\end{tabular}
\caption{Proxemic Zone Parameters}
\label{tab:proxemic_params}
\end{table}

\paragraph{Reward and Proxemic Parameters} Table~\ref{tab:proxemic_params} summarizes the proxemic-zone parameters used in Eq.~\ref{eq:1}. The Gaussian-mixture means $\mu_k$ define representative radii for the intimate, personal, social, and public zones, while the standard deviations $\sigma_k$ are chosen to produce smooth overlap between adjacent zones. $A_k$ decrease with distance to reflect stronger rewards for near-human intrusions than for far-field presence. The robot kernel width is set proportionally to the sensing radius, $\sigma_r = R_{\mathrm{FOV}}/1.5$, to provide a smooth but locally concentrated aggregation over the robot's FOV, and the proxemic weighting coefficient is fixed at $\lambda=15.0$ (see Sec.~\ref{sec:lambda} for ablations). Terminal reward parameters are $r_c=-10.0$ and $r_s=10.0$, which keep collision and reaching goal outcomes salient relative to dense shaping.

\subsection{Metrics}
Following the evaluation guidelines for social robot navigation in Francis et al. \cite{francis2025principles}, we report two complementary sets of metrics (Table~\ref{tab:metrics}): navigation metrics, which capture task effectiveness and physical safety, and social metrics, which quantify robot's human-aware behavior via proxemic compliance and motion smoothness. Social metrics are computed over successful episodes to isolate the human-comfort of navigation from task feasibility. Collisions and timeouts correspond to qualitatively different failure modes that (a) are already penalized and reported via SR/CR/TO, and (b) would otherwise dominate social measures in a non-informative way. In particular, failure episodes are often shorter and end at a terminal event, which distorts MD, SC and TTC statistics. This protocol provides a principled decomposition: navigation metrics evaluate safety and effectiveness, while social metrics evaluate compliance and comfort given that the robot actually reaches the goal.

\begin{table}[!h]
\centering
\setlength{\tabcolsep}{4pt}
\renewcommand{\arraystretch}{1.05}
\scriptsize

\subfloat[Navigation metrics.]{
\begin{tabular}{@{}p{3.1cm} p{4.2cm}@{}}
\hline
\textbf{Metric} & \textbf{Definition} \\
\hline
Success rate (SR) & Ratio of episodes in which the robot reaches its goal. \\
Collision rate (CR) & Proportion of episodes in which the robot collides with a human. \\
Timeout rate (TO) & Proportion of episodes that terminate before reaching the goal. \\
Travel length (TL) & Mean path length traveled by the robot across test episodes. \\
Travel time (TT) & Mean trajectory duration from start to goal across test episodes. \\
\hline
\end{tabular}
}\par\vspace{0.6ex}

\subfloat[Social metrics.]{
\begin{tabular}{@{}p{3.1cm} p{4.2cm}@{}}
\hline
\textbf{Metric} & \textbf{Definition} \\
\hline
Social compliance (SC) & Percentage of time the robot remains within a proxemic zone (sometimes referred to as the intrusion ratio). We report intimate ($d \le 0.25\,\mathrm{m}$) and personal ($d \le 0.45\,\mathrm{m}$) zones. \\
Minimum distance (MD) & Minimum distance to a human in a given episode. \\
Time to collision (TTC) & Time until collision with the closest human, assuming constant relative motion. \\
Jerk (JE) &  Mean magnitude of the derivative of acceleration along the episode. \\
\hline
\end{tabular}
}

\caption{Evaluation metrics.}\label{tab:metrics}
\end{table}

\subsection{Baselines and Reward Models}
\label{sec:rewards}
We evaluate our proposed reward model by integrating it into two DRL social navigation baselines, AttnGraph \cite{liu2023attngraph} and MultiSoc \cite{multisoc}, under the same training setting. The reward structure in Eq.~\ref{eq:6} is kept fixed, and only the social interaction term $r_{\text{soc},t}$ is changed across reward variants: 
\begin{itemize}
    \item \textbf{Base:}  $r_{\text{soc},t}=0$, i.e., no human-aware reward term is used. The agent is therefore trained only with the terminal and the base navigation rewards in Eq.~\ref{eq:6}.
    \item \textbf{Distance-based reward ($R_d$):} $r_{\text{soc},t}=r_{d,t}$, using the distance-based reward model proposed by Nishimura et al.~\cite{Nishimura2020L2B}. We use the hyperparameters reported in the original paper, revalidated in our experimental setup.
    \item \textbf{Velocity-based reward ($R_v$):} $r_{\text{soc},t}=r_{v,t}$, using the relative-velocity reward model proposed by Maddumage et al.~\cite{maddumage2025relative}. We use the hyperparameters reported in the original paper, revalidated in our experimental setup.
    \item \textbf{Proxemic-based reward ($R_p$):} $r_{\text{soc},t}=r_{\text{prox},t}$, i.e., the proposed proxemics-based reward model (Eq.~\ref{eq:5}).
\end{itemize}

We use PPO \cite{ppo} with a learning rate of $5\times10^{-5}$, $\gamma = 0.99$, $\lambda_{GAE} =0.95$, PPO clipping parameter$= 0.07$, 2 mini-batches, and 3 epochs per update. Each method is trained for $10^7$ timesteps with an episode length of $100$. Training is repeated over three random seeds. A single run with our reward model takes about 2.7 hours (vs. 2 hours without it). Since the reward is only used during training, inference incurs no additional overhead. For evaluation, all models are evaluated on the same randomized episodes. We report the mean and the standard deviation of the proposed metrics across the trained models over 100 episodes.

\begin{table*}[!h]
\centering
\scriptsize
\setlength{\tabcolsep}{3pt}
\renewcommand{\arraystretch}{1.0}

\begin{subtable}{\textwidth}
\centering
\begin{tabular}{@{}l l c c c c c | c c c c c@{}}
\hline
\multirow{2}{*}{\textbf{Method}} 
& \multirow{2}{*}{\textbf{Reward}} 
& \multicolumn{5}{c|}{\textbf{Navigation metrics}} 
& \multicolumn{5}{c}{\textbf{Social metrics}} \\
\cline{3-12}
& 
& \textbf{SR$\uparrow$} 
& \textbf{CR$\downarrow$} 
& \textbf{TO$\downarrow$}
& \textbf{TL$\downarrow$} 
& \textbf{TT$\downarrow$} 
& \textbf{MD$\uparrow$}
& \textbf{SC$_{0.25}\downarrow$}
& \textbf{SC$_{0.45}\downarrow$}
& \textbf{TTC$\uparrow$} 
& \textbf{JE$\downarrow$} \\
\hline

\multirow{4}{*}{AttnGraph}
& Base & $0.83_{(0.01)}$ & $0.13_{(0.02)}$ & $0.04_{(0.01)}$ & $11.78_{(0.21)}$ & $11.54_{(0.38)}$ & $0.48_{(0.02)}$ & $3.78_{(0.17)}$ & $11.08_{(0.28)}$ & $1.83_{(0.03)}$ & $3.05_{(0.07)}$ \\
& $R_d$ & $0.81_{(0.06)}$ & $0.16_{(0.08)}$ & $0.03_{(0.02)}$ & $\mathbf{11.24_{(0.21)}}$ & $\mathbf{10.96_{(0.71)}}$ & $0.53_{(0.02)}$ & $2.40_{(1.34)}$ & $8.84_{(2.50)}$ & $1.86_{(0.12)}$ & $3.33_{(0.53)}$ \\
& $R_v$ & $0.86_{(0.03)}$ & $0.13_{(0.03)}$ & $\mathbf{0.01_{(0.01)}}$ & $11.73_{(0.37)}$ & $11.33_{(0.41)}$ & $0.55_{(0.02)}$ & $2.03_{(0.24)}$ & $7.28_{(0.92)}$ & $1.69_{(0.04)}$ & $3.19_{(0.19)}$ \\
& $R_p$ & $\mathbf{0.87_{(0.01)}}$ & $\mathbf{0.12_{(0.01)}}$ & $\mathbf{0.01_{(0.02)}}$ & $12.56_{(0.02)}$ & $12.00_{(0.16)}$ & $\mathbf{0.65_{(0.02)}}$ & $\mathbf{1.42_{(0.32)}}$ & $\mathbf{6.55_{(0.34)}}$ & $\mathbf{2.38_{(0.14)}}$ & $\mathbf{2.47_{(0.19)}}$ \\
\hline

\multirow{4}{*}{MultiSoc}
& Base & $0.88_{(0.01)}$ & $0.11_{(0.01)}$ & $\mathbf{0.01_{(0.01)}}$ & $\mathbf{10.90_{(0.09)}}$ & $\mathbf{10.71_{(0.07)}}$ & $0.48_{(0.00)}$ & $3.51_{(0.21)}$ & $12.51_{(1.16)}$ & $1.75_{(0.11)}$ & $3.27_{(0.27)}$ \\
& $R_d$ & $0.81_{(0.03)}$ & $0.16_{(0.04)}$ & $0.03_{(0.01)}$ & $11.40_{(0.22)}$ & $10.72_{(0.62)}$ & $0.46_{(0.01)}$ & $3.20_{(0.25)}$ & $11.94_{(0.73)}$ & $1.75_{(0.12)}$ & $3.65_{(0.26)}$ \\
& $R_v$ & $0.85_{(0.03)}$ & $0.13_{(0.04)}$ & $0.02_{(0.01)}$ & $11.42_{(0.17)}$ & $10.81_{(0.45)}$ & $0.55_{(0.04)}$ & $3.73_{(0.75)}$ & $13.88_{(1.09)}$ & $1.94_{(0.12)}$ & $3.03_{(0.52)}$ \\
& $R_p$ & $\mathbf{0.89_{(0.03)}}$ & $\mathbf{0.10_{(0.01)}}$ & $\mathbf{0.01_{(0.02)}}$& $11.85_{(0.25)}$ & $11.66_{(0.31)}$ & $\mathbf{0.58_{(0.01)}}$ & $\mathbf{2.32_{(0.41)}}$ & $\mathbf{9.20_{(0.69)}}$ & $\mathbf{2.19_{(0.23)}}$ & $\mathbf{2.73_{(0.25)}}$ \\
\hline
\end{tabular}
\caption{Circle crossing scenario}
\label{tab:results_circle}
\end{subtable}

\vspace{0.4pt}

\begin{subtable}{\textwidth}
\centering
\begin{tabular}{@{}l l c c c c c | c c c c c@{}}
\hline
\multirow{2}{*}{\textbf{Method}} 
& \multirow{2}{*}{\textbf{Reward}} 
& \multicolumn{5}{c|}{\textbf{Navigation metrics}} 
& \multicolumn{5}{c}{\textbf{Social metrics}} \\
\cline{3-12}
& 
& \textbf{SR$\uparrow$} 
& \textbf{CR$\downarrow$} 
& \textbf{TO$\downarrow$}
& \textbf{TL$\downarrow$} 
& \textbf{TT$\downarrow$} 
& \textbf{MD$\uparrow$}
& \textbf{SC$_{0.25}\downarrow$}
& \textbf{SC$_{0.45}\downarrow$}
& \textbf{TTC$\uparrow$} 
& \textbf{JE$\downarrow$} \\
\hline

\multirow{4}{*}{AttnGraph}
& Base & $0.72_{(0.01)}$ & $0.27_{(0.02)}$ & $\mathbf{0.01_{(0.01)}}$ & ${15.35_{(0.21)}}$ & ${13.37_{(0.28)}}$ & $0.16_{(0.01)}$ & $6.39_{(0.06)}$ & $10.90_{(0.24)}$ & $0.79_{(0.11)}$ & $2.15_{(0.03)}$ \\

& $R_d$ & $0.67_{(0.09)}$ & $0.31_{(0.09)}$ & $\mathbf{0.01_{(0.01)}}$ & $15.53_{(0.82)}$ & $13.48_{(1.02)}$ & $0.34_{(0.08)}$ & $2.79_{(0.61)}$ & $7.86_{(2.16)}$ & $1.27_{(0.54)}$ & $2.10_{(0.32)}$ \\

& $R_v$  & $0.59_{(0.03)}$ & $0.37_{(0.02)}$ & ${0.03_{(0.03)}}$ & $\mathbf{15.16_{(0.14)}}$ & $\mathbf{12.79_{(0.24)}}$ & $0.35_{(0.08)}$ & $2.64_{(0.08)}$ & $6.61_{(0.81)}$ & $1.30_{(0.14)}$ & $2.14_{(0.19)}$ \\

& $R_p$ & $\mathbf{0.74_{(0.02)}}$ & $\mathbf{0.23_{(0.02)}}$ & ${0.03_{(0.01)}}$ & $16.23_{(0.13)}$ & $14.41_{(0.28)}$ & $\mathbf{0.67_{(0.01)}}$ & $\mathbf{2.41}_{(0.28)}$ & $\mathbf{5.49}_{(0.53)}$ & $\mathbf{1.96_{(0.10)}}$ & $\mathbf{2.00_{(0.03)}}$ \\

\hline

\multirow{4}{*}{MultiSoc}
& Base & $0.60_{(0.05)}$ & $0.38_{(0.05)}$ & $0.02_{(0.00)}$ & $\mathbf{14.59}_{(0.35)}$ & $\mathbf{12.38}_{(0.58)}$ & $0.19_{(0.03)}$ & $5.13_{(0.32)}$ & $10.08_{(0.53)}$ & $0.77_{(0.12)}$ & $2.50_{(0.11)}$ \\

& $R_d$ & $0.66_{(0.05)}$ & $0.33_{(0.03)}$ & $0.01_{(0.02)}$ & $14.77_{(0.42)}$ & $12.61_{(0.48)}$ & $0.35_{(0.11)}$ & $4.63_{(1.01)}$ & $8.70_{(1.11)}$ & $1.71_{(0.25)}$ & $1.91_{(0.15)}$ \\

& $R_v$ & $0.77_{(0.02)}$ & $\mathbf{0.19}_{(0.00)}$ & $0.04_{(0.02)}$ & $16.57_{(0.03)}$ & $15.11_{(0.02)}$ & $0.61_{(0.01)}$ & $4.86_{(0.70)}$ & $8.80_{(0.74)}$ & $1.88_{(0.03)}$ & $1.87_{(0.14)}$ \\

& $R_p$ & $\mathbf{0.80}_{(0.05)}$ & $0.20_{(0.05)}$ & $\mathbf{0.00}_{(0.01)}$ & $16.20_{(0.48)}$ & $14.52_{(0.63)}$ & $\mathbf{0.71}_{(0.09)}$ & $\mathbf{4.22_{(0.79)}}$ & $\mathbf{8.19_{(0.33)}}$ & $\mathbf{2.13}_{(0.07)}$ & $\mathbf{1.81_{(0.02)}}$ \\

\hline
\end{tabular}
\caption{Corridor scenario}
\label{tab:results_corridor}
\end{subtable}

\caption{Performance comparison of AttnGraph and MultiSoc under four reward variants across evaluation scenarios.}
\label{tab:results}
\end{table*}

\section{Experimental results}
\label{sec:experimental_results}

\subsection{Comparative analysis} 

\begin{figure*}[h]
    \centering
    \begin{subfigure}[c]{0.45\columnwidth}
        \includegraphics[width=\textwidth]{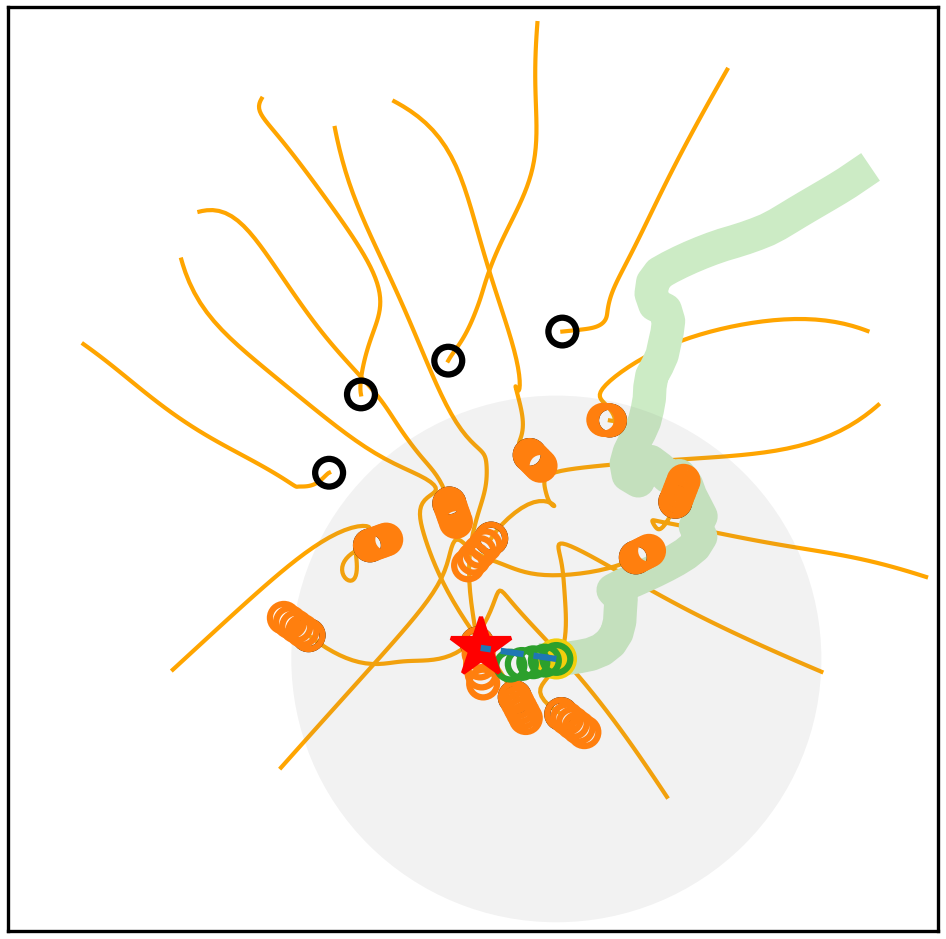}
         \caption{}
    \end{subfigure}
    \begin{subfigure}[c]{0.45\columnwidth}
        \includegraphics[width=\textwidth]{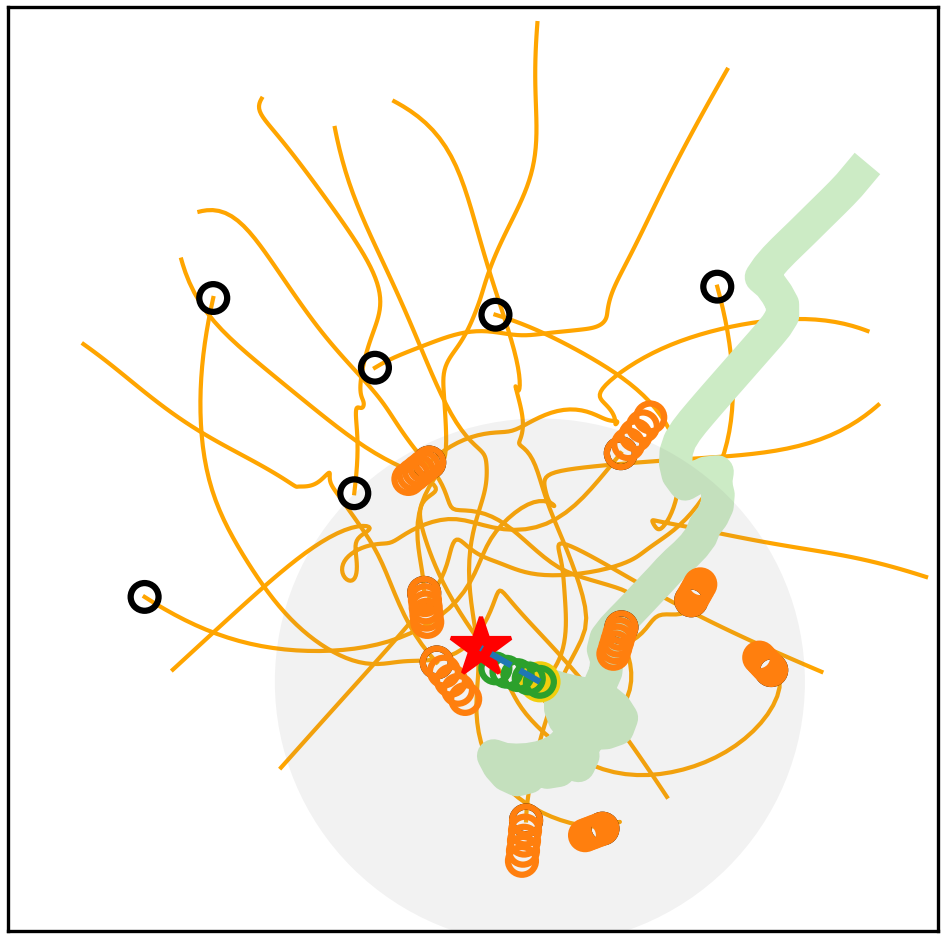}
         \caption{}
    \end{subfigure}
    \begin{subfigure}[c]{0.45\columnwidth}
    \includegraphics[width=\textwidth]{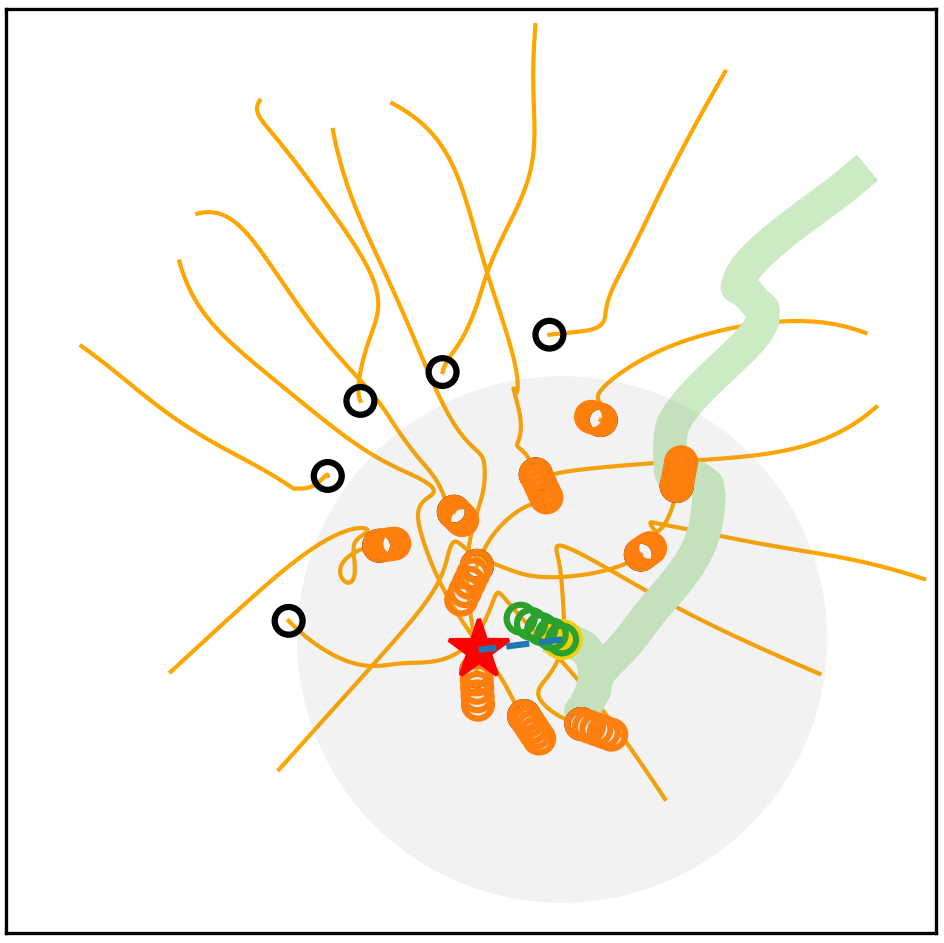}
         \caption{}
    \end{subfigure}
    \begin{subfigure}[c]{0.45\columnwidth}
        \includegraphics[width=\textwidth]{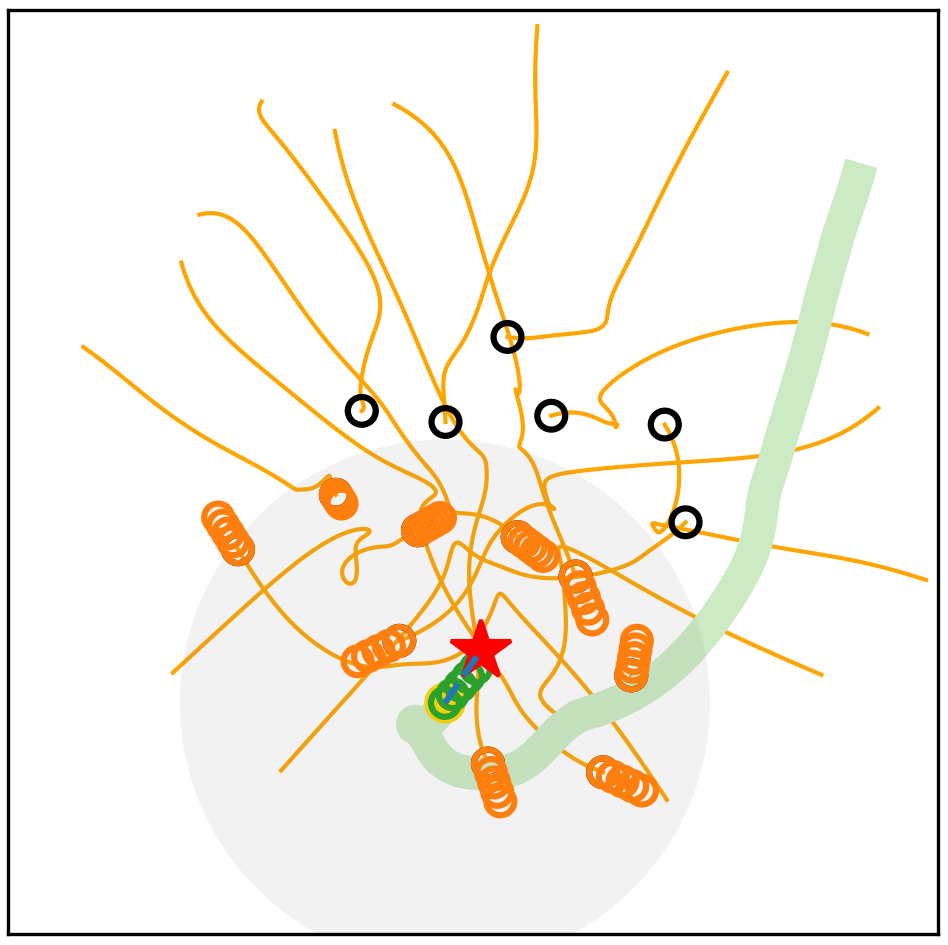}
         \caption{}
    \end{subfigure}
    \\[0.5em]
    \begin{subfigure}[c]{0.45\columnwidth}
        \includegraphics[width=\textwidth]{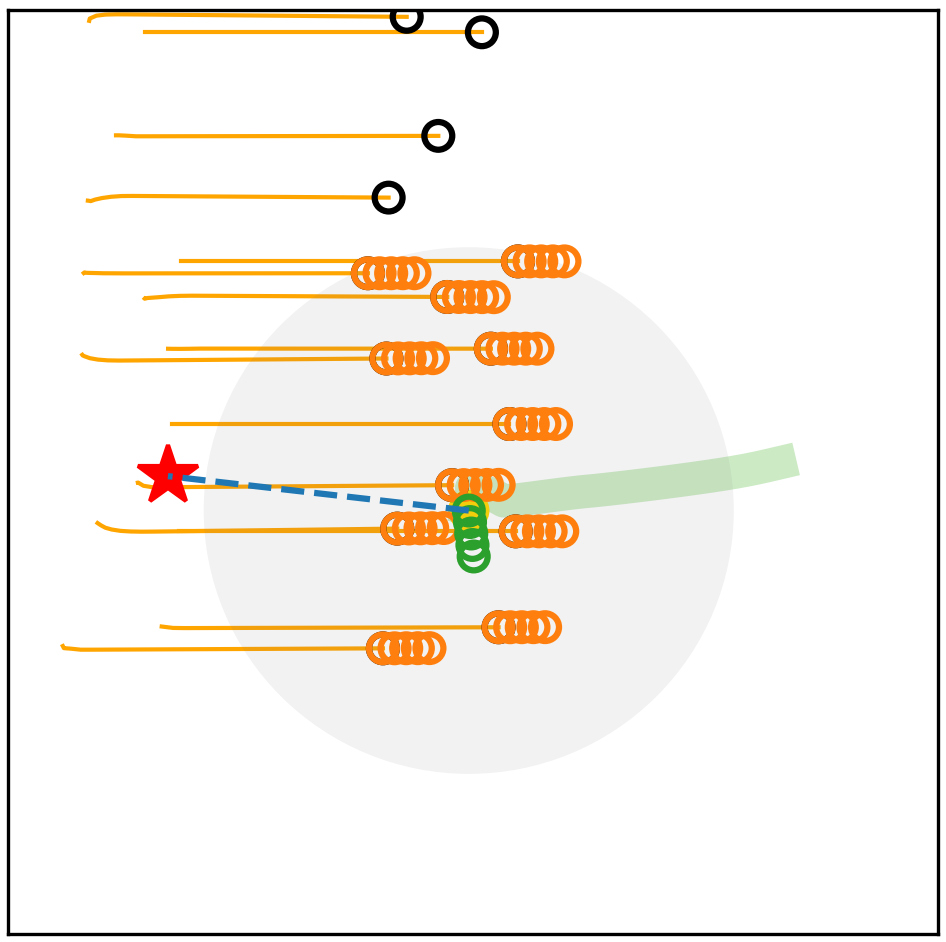}
         \caption{}    
    \end{subfigure}
    \begin{subfigure}[c]{0.45\columnwidth}
        \includegraphics[width=\textwidth]{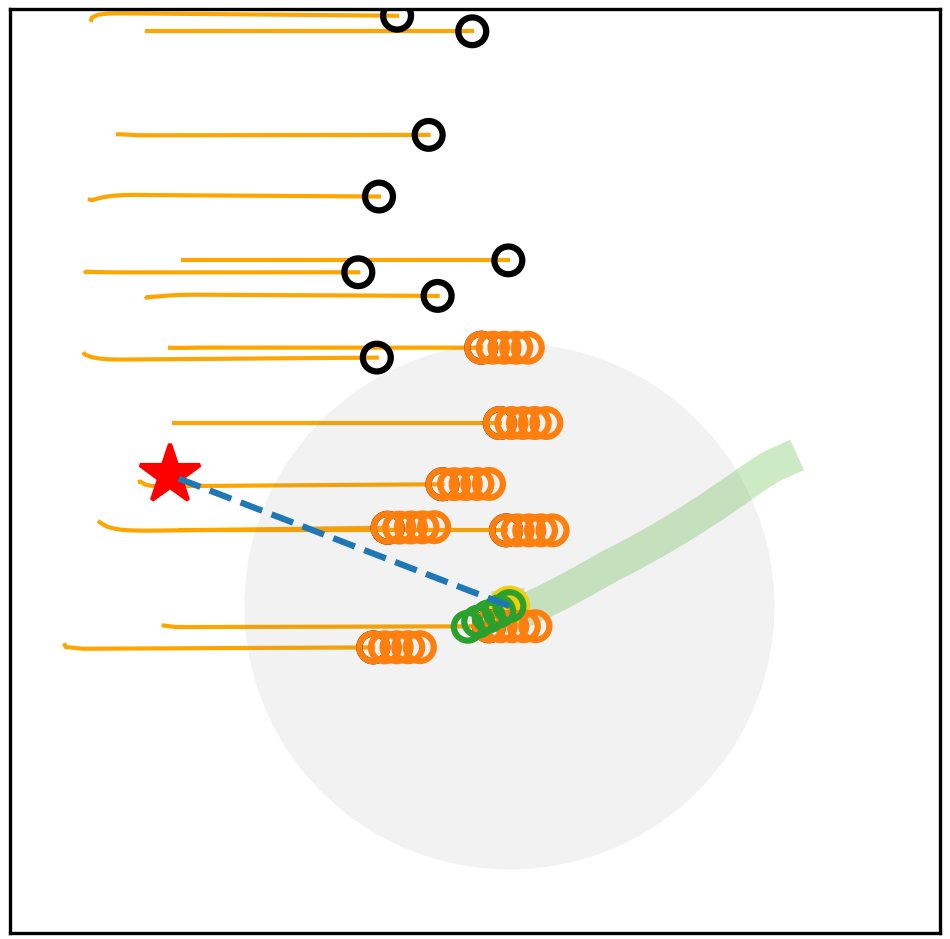}
         \caption{}
    \end{subfigure}
    \begin{subfigure}[c]{0.45\columnwidth}
        \includegraphics[width=\textwidth]{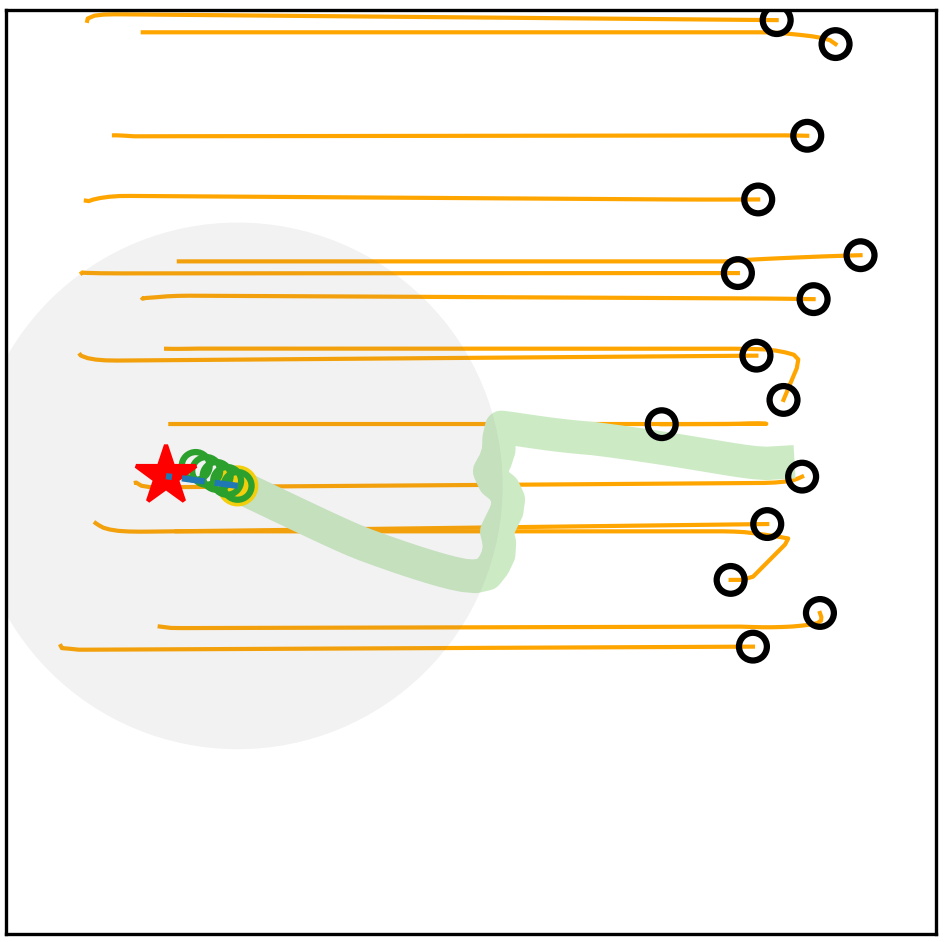}
         \caption{}
    \end{subfigure}
    \begin{subfigure}[c]{0.45\columnwidth}
        \includegraphics[width=\textwidth]{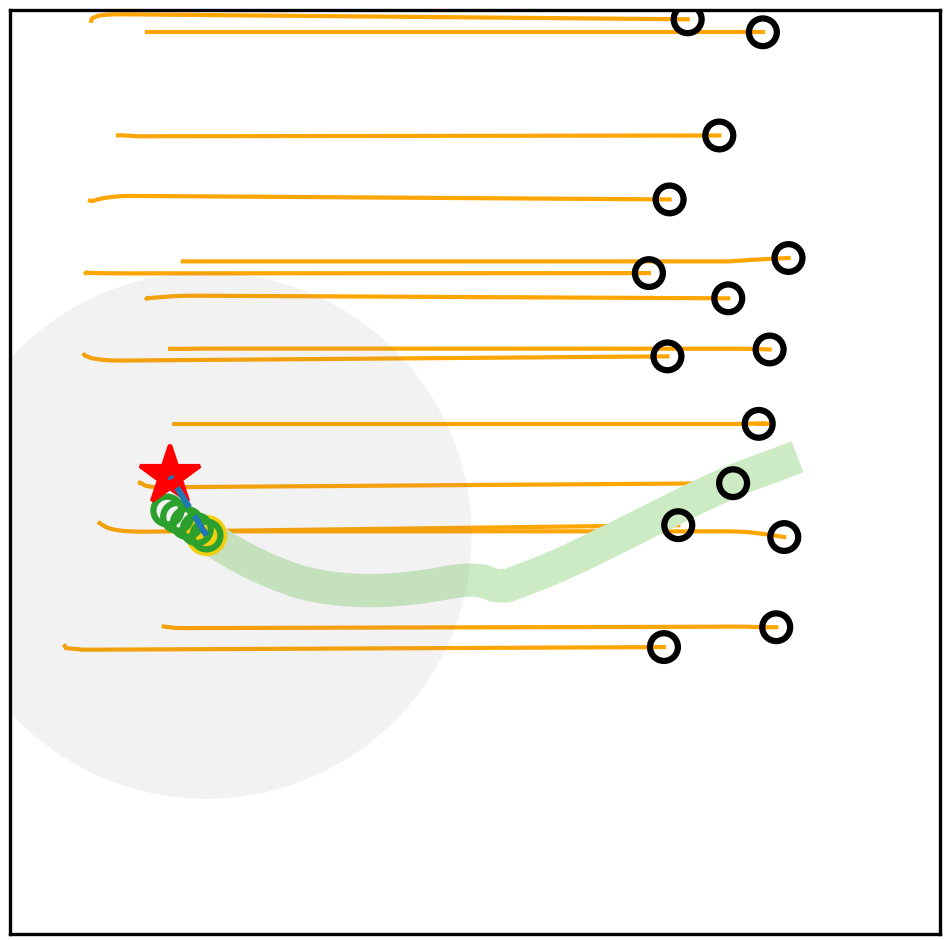}
         \caption{}
    \end{subfigure}

    \caption {Sample trajectories for AttnGraph in the Simple Circle scenario (top row) and MultiSoc in the Corridor scenario (bottom row) under the compared reward models. Detected humans are shown in orange, undetected humans in white, and the robot in green. Subfigures (a)--(d) and (e)--(h) correspond to the Base, $R_d$, $R_v$, and $R_p$ settings, respectively.}
    \label{fig:simulation_results}
\end{figure*}

Table~\ref{tab:results} shows results for both the Circle Crossing and Corridor scenarios across the compared reward models and DRL navigation methods. Across both settings, the proposed reward model $R_p$ yields the strongest overall gains on social metrics while preserving most of the navigation performance. From a social navigation perspective, this suggests that explicitly structuring the reward around proxemic zones aligns the learning objective with maintaining comfortable minimum distances, rather than merely avoiding hard collisions. This induces policies that keep larger separations from nearby humans and spend less time within intimate zones. The corresponding increase in anticipatory safety, reflected by longer TTC, suggests that the robot commits to avoidance earlier and selects trajectories that resolve future conflicts before they become imminent. In addition, JE is lower for both MultiSoc and AttnGraph, indicating smoother motion, which suggests that proxemic shaping encourages earlier and more gradual trajectory adjustments around humans, reducing abrupt reactive maneuvers. From a DRL standpoint, these effects are consistent with reward shaping that provides a smoother and more informative gradient: rewarding graded proxemic zones over the robot's FOV creates dense feedback across a wide range of interactions, reducing reliance on sparse terminal signals, which can make socially acceptable behaviors harder to learn. In contrast, $R_d$ can tolerate frequent intrusions, while $R_v$ emphasizes imminent collision risk without explicitly encoding socially meaningful behavior. Qualitatively, Figure~\ref{fig:simulation_results} showcases the trends in Table~\ref{tab:results}. For both MultiSoc and AttnGraph, policies trained with $R_p$ consistently select wider, more peripheral passing arcs, avoiding dense intersections of crowd flow. By comparison, $R_d$ and $R_v$ produce more central, closer pass-bys and more reactive trajectory changes near pedestrians.

\subsection{Ablations}
In this subsection, we present ablation results. All reported results are on the circle crossing scenario.
\subsubsection{Human scalability}
\label{sec:human}
\begin{figure}[!h]
    \centering

\includegraphics[width=\columnwidth]{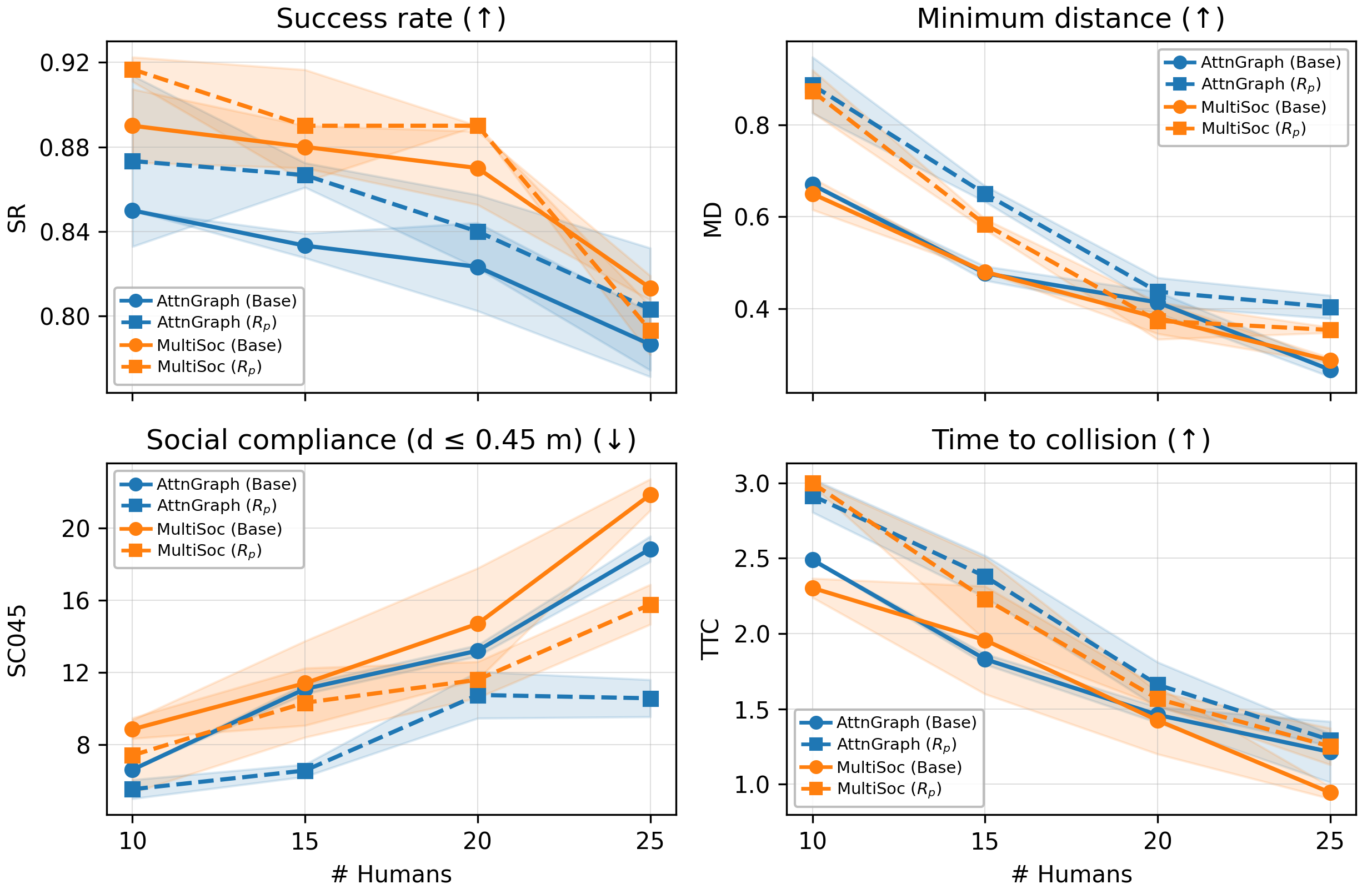}

\caption{Performance as a function of crowd size . We report SR, MD, SC$_{0.45}$ and TTC. Curves show the mean and standard deviation.}
    \label{fig:scalability}
\end{figure}

Figure~\ref{fig:scalability} reports how task performance and social comfort scale with crowd density, and how the proposed proxemic reward $R_p$ shifts the learned behavior. As the number of humans increases, both AttnGraph and MultiSoc exhibit the expected degradation in performance. Success rate decreases while the interaction margins shrink, reflected by reduced MD and shorter TTC. This trend is consistent with the adversarial-crowd setting and a fixed-horizon objective, where higher density compresses feasible free space and increases the frequency of close encounters. Incorporating $R_p$ produces a consistent and interpretable behavioral shift for both AttnGraph and MultiSoc. The robot maintains larger clearance from pedestrians, yielding higher MD and TTC, and it spends less time within the personal zone, reflected by the  $\mathrm{SC}_{0.45}$. These effects show that $R_p$ provides an informative signal which biases the policy toward socially compliant interaction geometries (e.g., wider passing arcs and more conservative gap selection) rather than merely optimizing purely for reaching goal coordinates. 


 \subsubsection{Weighting sensitivity}
 \label{sec:lambda}
\begin{figure}[!h]
    \centering

\includegraphics[width=\columnwidth]{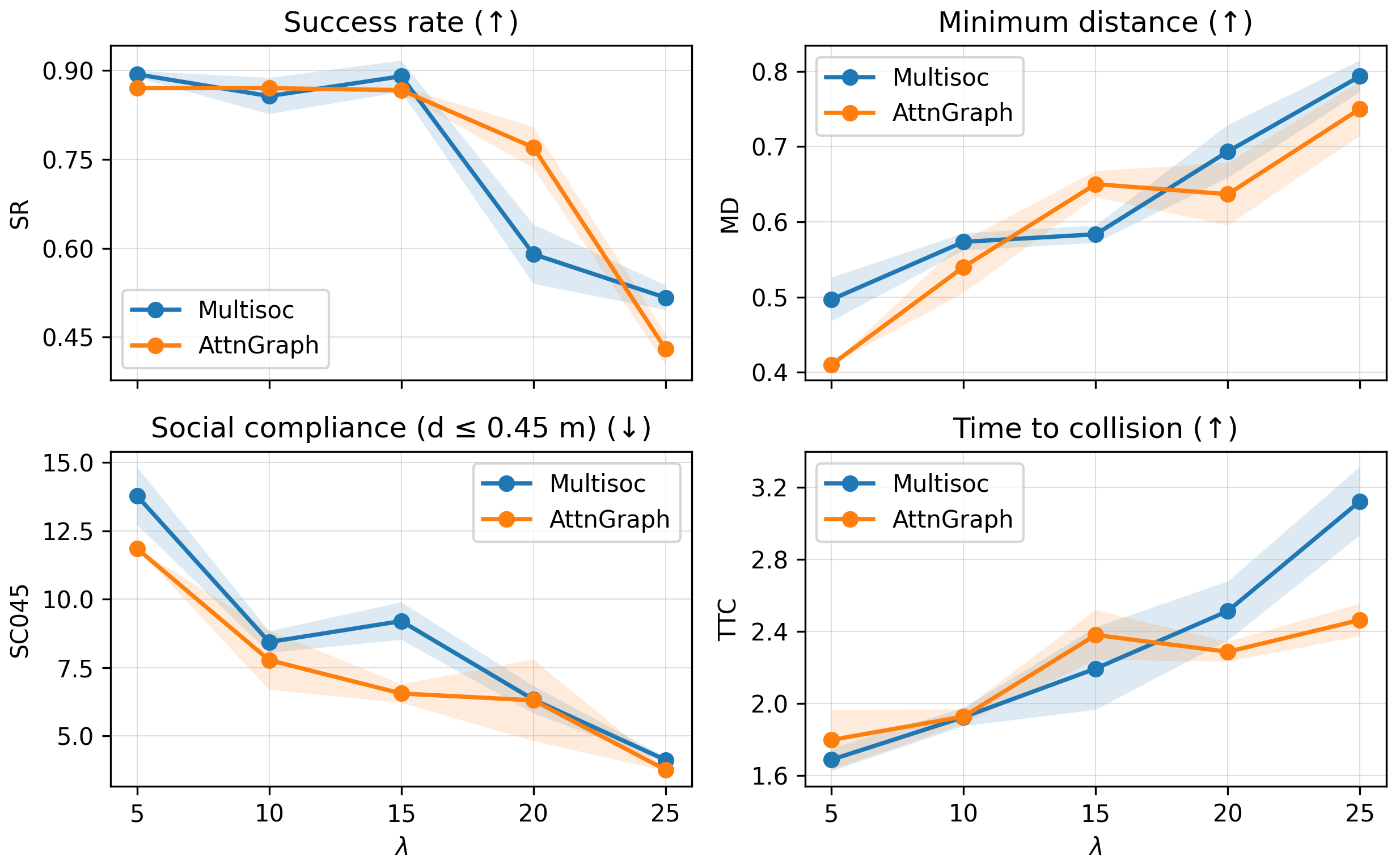}

\caption{Effect of the proxemic reward weight $\lambda$ on performance. We report SR, MD, SC$_{0.45}$ and TTC as a function of $\lambda$. Curves show the mean and standard deviation. }
\label{fig:lambda}
\end{figure}

Figure~\ref{fig:lambda} analyzes the sensitivity of the proxemic reward weighting coefficient $\lambda$ (Eq.~\ref{eq:5}) and highlights the resulting trade-off between navigation completion and socially compliant interaction. For both AttnGraph and MultiSoc, increasing $\lambda$ pushes the model to respect proxemic costs, which consistently expands the robot’s interaction margin: the MD increases and the predicted TTC grows, while  SC$_{0.45}$ decreases, indicating fewer and shorter incursions into human proximity. These trends are characteristic of reward shaping that is well-aligned with social-navigation objectives, as the policy internalizes proxemic preferences by selecting wider passing geometries and earlier yielding behaviors rather than reacting late in close proximity. At the same time, overly large $\lambda$ reduces the success rate, suggesting that an excessively dominant proxemic term can over-regularize the policy and impede  the navigation progress. Indeed, the robot becomes more conservative, avoids narrow but feasible gaps. The overall pattern indicates an intermediate operating regime in which $\lambda$ is strong enough to meaningfully improve social compliance margins without overly sacrificing efficiency, where $\lambda$ acts as a knob to balance between navigation efficiency and social compliance.

\begin{table*}[!h]
\centering
\scriptsize
\renewcommand{\arraystretch}{1.2}
\resizebox{\textwidth}{!}{
\begin{tabular}{@{}l l r r r r r | r r r r r r r@{}}
\hline
\multirow{2}{*}{\textbf{Method}}
& \multirow{2}{*}{\textbf{Reward}}
& \multicolumn{5}{c|}{\textbf{Navigation metrics}}
& \multicolumn{7}{c}{\textbf{Social metrics}} \\
\cline{3-14}
&
& \textbf{SR$\uparrow$}
& \textbf{CR$\downarrow$}
& \textbf{TO$\downarrow$}
& \textbf{TL$\downarrow$}
& \textbf{TT$\downarrow$}
& \textbf{MD$\uparrow$}
& \textbf{SC$_{0.25}\downarrow$}
& \textbf{SC$_{0.45}\downarrow$}
& \textbf{SC$_{0.45}^{F}$}
& \textbf{SC$_{0.45}^{B}$}
& \textbf{TTC$\uparrow$}
& \textbf{JE$\downarrow$} \\
\hline
\multirow{2}{*}{AttnGraph}
& $R_p$-Ani.
& $0.85_{(0.01)}$ & $0.15_{(0.02)}$ & $\mathbf{0.00_{(0.01)}}$ & $\mathbf{11.82_{(0.57)}}$ & $\mathbf{11.18_{(0.50)}}$
& $0.57_{(0.03)}$ & $4.08_{(0.87)}$ & $11.65_{(1.92)}$ &  $52.11_{(3.51)}$  & $47.89_{(3.53)}$ & $2.07_{(0.27)}$ & $2.70_{(0.01)}$ \\
& $R_p$-Iso.
& $\mathbf{0.87_{(0.01)}}$ & $\mathbf{0.12_{(0.01)}}$ & $0.01_{(0.02)}$ & $12.56_{(0.02)}$ & $12.00_{(0.16)}$
& $\mathbf{0.65_{(0.02)}}$ & $\mathbf{1.42_{(0.32)}}$ & $\mathbf{6.55_{(0.34)}}$ & $51.40_{(2.31)}$ & $48.60_{(2.31)}$ & $\mathbf{2.38_{(0.14)}}$ & $\mathbf{2.47_{(0.19)}}$ \\
\hline
\multirow{2}{*}{MultiSoc}
& $R_p$-Ani.
& $\mathbf{0.90_{(0.01)}}$ & $\mathbf{0.10_{(0.00)}}$ & $\mathbf{0.00_{(0.01)}}$ & $\mathbf{11.23_{(0.35)}}$ & $\mathbf{10.93_{(0.34)}}$
& $0.51_{(0.01)}$ & $5.06_{(0.14)}$ & $13.78_{(4.11)}$ & $47.54_{(1.40)}$ & $52.46_{(1.40)}$ & $1.71_{(0.03)}$ & $3.48_{(0.40)}$ \\
& $R_p$-Iso.
& $0.89_{(0.03)}$ & $0.10_{(0.01)}$ & $0.01_{(0.02)}$ & $11.85_{(0.25)}$ & $11.66_{(0.31)}$
& $\mathbf{0.58_{(0.01)}}$ & $\mathbf{2.32_{(0.41)}}$ & $\mathbf{9.20_{(0.69)}}$ & $47.74_{(1.46)}$ & $52.26_{(1.46)}$ & $\mathbf{2.19_{(0.23)}}$ & $\mathbf{2.73_{(0.25)}}$ \\
\hline
\end{tabular}}
\caption{Comparison between anisotropic and isotropic proxemic reward formulations.}
\label{tab:isotropic_vs_anisotropic}
\end{table*}

\subsubsection{Field of view}
\label{sec:fov}
\begin{figure}[!h]
\centering
\includegraphics[width=\columnwidth]{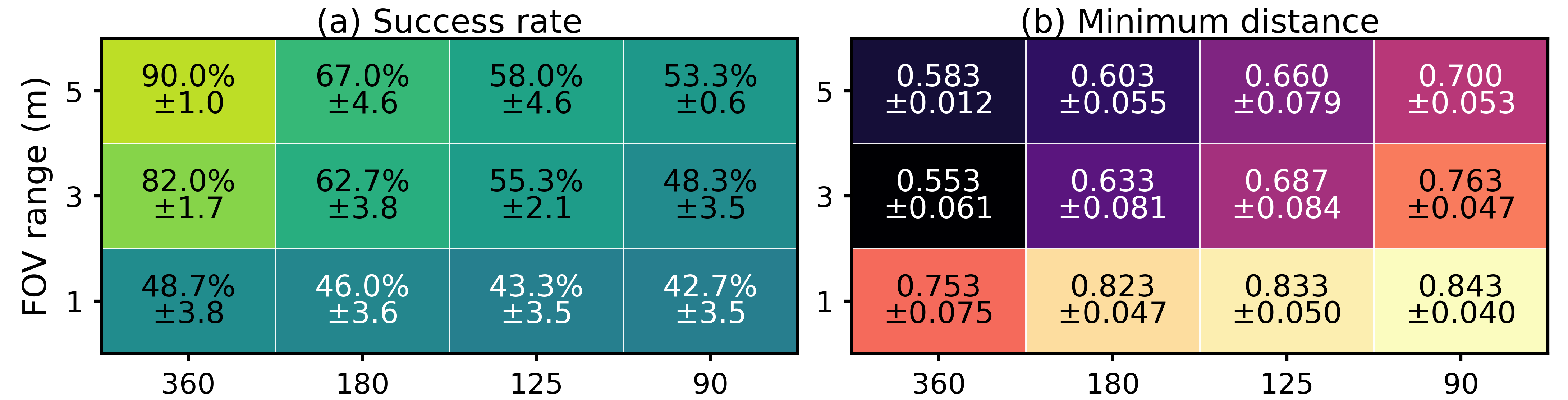}
\includegraphics[width=\columnwidth]{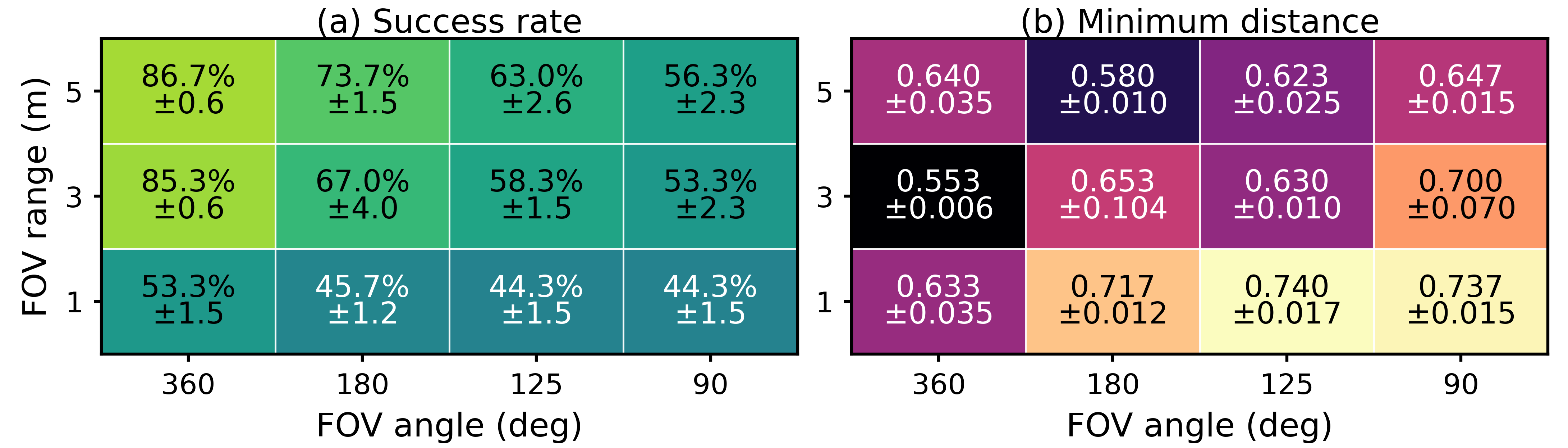}
\caption{Effect of FOV parameters on performance. Heatmaps report (a) SR and (b) MD as functions of FOV angle and range for MultiSoc (up) and AttnGraph (bottom).}
\label{fig:fov_ablation}
\end{figure}

Figure~\ref{fig:fov_ablation} shows the effect kernel $\kappa$ parameters $R_{\mathrm{FOV}}$ and FOV on navigation performance for both MultiSoc and AttnGraph. Across both methods, the SR heatmaps reveal a consistent pattern. Using larger FOV generally improves task SR, with the strongest gains obtained under wider angular coverage and longer sensing range. This suggests that socially compliant navigation benefits from richer local situational awareness, as broader perception enables earlier anticipation and less reactive conflict resolution. In contrast, narrow, short range FOVs consistently degrade SR, likely due to reduced observability of nearby crowd dynamics and delayed responses to emerging conflicts. The MD heatmaps show a complementary trade-off. In several weak configurations, larger MD coincides with lower SR, suggesting the increased clearance reflects easier, less dense interactions rather than better social behavior. Conversely, broader FOV configurations often achieve higher SR despite lower MD, suggesting improved coordination in denser interactions rather than simply more aggressive motion. Overall, the results indicate that FOV design is an important factor in DRL social navigation, and that wide-angle FOV with sufficient range provides a more informative observation space for learning robust human-aware behavior.

\subsubsection{Gaussian shape}
Table~\ref{tab:isotropic_vs_anisotropic} compares two proxemic reward formulations: an anisotropic formulation, whose cost field is aligned with each pedestrian’s motion direction, and an isotropic formulation, which depends only on relative distance. The isotropic reward consistently performs better for both AttnGraph and MultiSoc across navigation and social metrics. We attribute this to DRL learning dynamics: isotropic shaping provides a smoother, denser, and direction-agnostic reward signal, yielding more reliable feedback for policy learning, whereas anisotropic fields introduce sharp directional discontinuities and rapidly varying penalties around pedestrian headings. Under stochastic exploration and frequent heading changes in dense crowds, this creates a higher-variance reward landscape that is harder to optimize. To test whether either formulation induces a directional passing preference, we additionally report front/back intrusion ratios, $\mathrm{SC}_{0.45}^{F}$ and $\mathrm{SC}_{0.45}^{B}$. Our results show that these ratios remain close to balanced for both anisotropic and isotropic variants, indicating no clear learned preference for front versus back-side intrusions. Thus, in our DRL setting, anisotropic shaping does not translate into a consistent directional avoidance strategy, in contrast to model/optimization-based social navigation, where anisotropic Gaussians are often preferred to exploit directional asymmetry.

\subsection{Discussion and limitations}
Across AttnGraph and MultiSoc, explicitly encoding proxemics in the reward consistently shifts behavior toward human-aware navigation, not just better collision avoidance. The proposed reward improves the social metrics while remaining competitive on task performance, supporting the idea of proxemics-based reward modelling as a continuous local comfort signal over the robot’s FOV. The scalability and $\lambda$-sensitivity studies further show that the method acts as a tunable knob on the efficiency/compliance trade-off: increasing the weight monotonically promotes safer, more socially compliant motion, but can reduce navigation efficiency in dense crowds as the policy becomes overly cautious. The isotropic vs. anisotropic ablation also suggests a key DRL specific effect: although anisotropic costs may help classical planners exploit directional asymmetry, the isotropic formulation provides a smoother, more robust learning signal under stochastic exploration and yields better overall performance. A key limitation is that reward shaping is a soft constraint: it encourages compliance but does not guarantee constraint satisfaction. A natural next step is to extend the proposed proxemics model within a constrained RL framework where strict adherence is enforced.

\section{Conclusion}
\label{sec:conclusion}
In this paper, we introduced a proxemics-informed reward model for deep reinforcement learning in human-aware navigation. The proposed model defines a continuous discomfort cost field from proxemic zones around nearby humans over the robot’s field of view. Using this cost as a training signal, the resulting policy encourages the robot to reduce perceived discomfort while maintaining task performance. Future work will address the highlighted limitation of reward-based constraint modeling, incorporate a human preference study, and focus on deploying the trained policies on a real robot.

\bibliographystyle{IEEEtran}

\bibliography{bibliography}






\end{document}